\documentclass[final]{cvpr}

\usepackage{times}
\usepackage{epsfig}
\usepackage{graphicx}
\usepackage{amsmath}
\usepackage{amssymb}
\usepackage[ruled,vlined]{algorithm2e}
\usepackage{bm}
\usepackage{booktabs}
\usepackage{multirow}
\usepackage{subfigure}
\usepackage{color}

\usepackage[pagebackref=true,breaklinks=true,colorlinks,bookmarks=false]{hyperref}

\def\cvprPaperID{3699} % *** Enter the CVPR Paper ID here
\def\confYear{CVPR 2021}

\begin{document}

%%%%%%%%% TITLE
\title{ Meta-Generating Deep Attentive Metric for Few-shot Classification}

\author{Lei Zhang, Fei Zhou, Wei Wei, Yanning Zhang\\
School of computer science, Northwestern Polytechnical University}
%\and

\maketitle

%%%%%%%%% ABSTRACT
\begin{abstract}
Learning to generate a task-aware base learner proves a promising direction to deal with few-shot learning (FSL) problem. Existing methods mainly focus on generating an embedding model utilized with a fixed metric (\eg, cosine distance) for nearest neighbour classification or directly generating a linear classier. However, due to the limited discriminative capacity of such a simple metric or classifier, these methods fail to generalize to challenging cases appropriately. To mitigate this problem, we present a novel deep metric meta-generation method that turns to an orthogonal direction, \ie, learning to adaptively generate a specific metric for a new FSL task based on the task description (\eg, a few labelled samples). In this study, we structure the metric using a three-layer deep attentive network that is flexible enough to produce a discriminative metric for each task. Moreover, different from existing methods that utilize an uni-modal weight distribution conditioned on labelled samples for network generation, the proposed meta-learner establishes a multi-modal weight distribution conditioned on cross-class sample pairs using a tailored variational autoencoder, which can separately capture the specific inter-class discrepancy statistics for each class and jointly embed the statistics for all classes into metric generation. By doing this, the generated metric can be appropriately adapted to a new FSL task with pleasing generalization performance. To demonstrate this, we test the proposed method on four benchmark FSL datasets and gain surprisingly obvious performance improvement over state-of-the-art competitors, especially in the challenging cases, \eg, {\textbf{improve the accuracy from 26.14\% to 46.69\% in the 20-way 1-shot task on miniImageNet, while improve the accuracy from 45.2\% to 68.72\% in the 5-way 1-shot task on FC100}}. {\textcolor{magenta}{Code is available: {\url{https://github.com/NWPUZhoufei/DAM.}}}}

\end{abstract}

%%%%%%%%% BODY TEXT
\section{Introduction}
Recently machine learning has achieved great success in a variety of computer vision tasks, such as image restoration~\cite{zhang2020unsupervised,laine2019high}, pedestrian detection~\cite{liu2019high,pang2019mask,song2020progressive}, object tracking~\cite{braso2020learning,voigtlaender2019mots}, image classification~\cite{sainath2013deep,he2016deep,huang2017densely} or video analysis~\cite{shi2019skeleton,kazakos2019epic} etc. This mainly profits from the widespread usage of the deep neural networks (DNNs)~\cite{lecun2015deep,sainath2013deep,he2016deep} that embed extensive learnable weights into deeply cascaded network architectures and show flexible fitting capacity in modelling high-dimensional non-linear relations. However, to properly train an DNN, it necessitates extensive labelled samples which cause expensive labour cost~\cite{lee2019meta,li2019lgm}. In contrast, human can learn a concept quickly based on one~\cite{lake2011one,fei2006one} or a few samples~\cite{tokmakov2019learning}, \eg, a child can easily generalize the concept of 'tiger' in a picture to a real one. Inspired by these two aspects, increasing effort~\cite{snell2017prototypical,mishra2017simple,munkhdalai2018rapid,sung2018learning,
oreshkin2018tadam,satorras2018few,bertinetto2018meta,lee2019meta,li2019lgm} has been made to investigate the few-shot learning (FSL) problem.

Profiting from the ability of fast adapting prior knowledge of existing FSL tasks to a new one, meta-learning technique~\cite{vilalta2002perspective,finn2017model} has underpinned much recent progress in FSL~\cite{mishra2017simple,munkhdalai2018rapid,
oreshkin2018tadam,bertinetto2018meta,lee2019meta,li2019lgm,feat_cvpr20,
Hu2020Empirical,zhang2020deepemd}. In general, a meta-learning framework consists of a base learner that solves the target task and a meta-learner that transfers prior knowledge from existing tasks to the target one. A promising meta-learning scheme proves developing a meta-learner that learns to generate a task-aware base learner based on the task description (\eg, a few labelled samples)~\cite{oreshkin2018tadam,Gidaris_2018_CVPR,li2019lgm,chen2020diversity,feat_cvpr20}. In this way, the meta-learner can be trained in an end-to-end manner and circumvents the complicated second-order gradient computation~\cite{finn2017model,ravi2017optimization,lee2019meta}. Moreover, the generated base learner can be directly deployed onto a new task without further optimization~\cite{finn2017model,bertinetto2018meta,lee2019meta}. Inspired by these, some works propose learning to generate an embedding module that is then utilized with a fixed metric (\eg, cosine distance) for nearest neighbour classification~\cite{lee2019meta,li2019lgm,feat_cvpr20} or directly generate a linear classier~\cite{chen2020diversity}. However, due to the limited discriminative capacity caused by their shallow and simple structure, such a metric based nearest neighbour classifier or the linear classier fail to appropriately generalize to the challenging FSL tasks.

To mitigate this problem, we present a deep metric meta-generation approach. Instead of generating an embedding module~\cite{lee2019meta,li2019lgm,feat_cvpr20} or a linear classifier~\cite{chen2020diversity}, we propose to learn to adaptively generate a specific deep metric for a new FSL task. Following this idea, we implement a meta-learner using a tailored variational autoencoder~\cite{kingma2019introduction,pu2016variational} which establishes a multi-modal weight distribution conditioned on cross-class sample pairs and then generates a deep metric network via sampling the weight distribution. In the multi-modal weight distribution, the statistics of the discrepancy between each class to the remaining ones are separately represented by a latent Gaussian distribution in a deep feature space. By doing this, the task related information (\eg, task-aware inter-class discrepancy) will be embedded into the multi-modal weight distribution and thus enables the generated metric to be adapted to a new FSL task. Moreover, in order to obtain a flexible discriminative metric for a new FZL task, we formulate the metric using a three-layer deep attentive network, where an element-wise attention module is utilized to highlight the discriminative information of the input data. With the generated task-aware deep attentive metric network, the proposed method shows pleasing generalization performance on four benchmark FSL datasets, especially in challenging cases.

In summary, the contribution of this study mainly comes from the following three aspects.

\begin{itemize}
  \item We present a novel meta-learning method that learns to generate a task-aware deep attentive metric for a new FSL task. To the best of our knowledge, this is the first attempt to do so in FSL domain. 
  \item We employ a tailored variational autoencoder as the meta-learner that can sufficiently exploit the task context for metric generation via establishing a multi-modal weight distribution conditioned on the cross-class sample pairs.
  \item We demonstrate the state-of-the-art performance on four FSL benchmark datasets.
\end{itemize}

\section{Related work}
In this part, we will briefly review three categories of existing meta-learning FSL methods related to this study.

\subsection{Metric based Method} Metric based method focuses on establishing a nearest neighbor classifier for a new FSL task using a fixed metric with the learned embedding modules or a learned metric. For example, Snell \etal~\cite{snell2017prototypical} propose to represent each class with the mean embeddings of the labelled samples belonging to this class and then utilize the cosine distance as the metric for classification. Vinyals \etal~\cite{vinyals2016matching} employ an LSTM network~\cite{malhotra2015long} to capture the task-aware context and then formulate an attention kernel based similarity metric for prediction. Sung \etal~\cite{sung2018learning} turn to learn a deep distance metric for FSL tasks. However, either the learned embedding module or the learned metric is still fixed for all FSL tasks, which fails to adaptively fit the characteristics of each task and thus limits the generalization performance. In this study, we propose to learn to generate a specific deep metric for each task and achieve better generalization performance.

\subsection{Gradient based Method} Gradient based method aims at developing a meta learner to learn a specific gradient based optimization algorithm used to train the base learner. For example, Ravi \etal~\cite{ravi2017optimization} develop an LSTM based meta learner to learn appropriate weight update specifically for the scenario where a set amount of updates will be made, while also learn a general initialization of the base learner that allows for fast convergence of training. Finn \etal~\cite{finn2017model} learn to initialize the base learner and fine-tune it using several steps of gradient descend. Li \etal~\cite{Li2017Meta} develop a meta-learner that initializes and adapts the base learner in just one step. Rusu \etal~\cite{rusu2018meta} learn a gradient based optimization based on the initialized weights sampled from a data-dependent low-dimensional weight representation. Recently, Lee et \etal~\cite{lee2019meta} learn to optimize feature embeddings that generalize well under a linear classifier for novel categories. However, these methods often require to compute the complicated second-order gradient of model weights which increases the training complexity of the meta learner. In contrast, the proposed meta learner can directly generate a viable base learner without further optimization, which enable training the meta leaner in an end-to-end manner.

\subsection{Generation based Method} Generation based method develops a meta-learner that learns to generate a base learner. For example, Li \etal~\cite{li2019lgm} propose learning to generate a task-aware embedding module and then utilize it with a cosine distance metric for classification. Qiao \etal~\cite{Qiao_2018_CVPR} utilize a adaptively generated top layer to adapt the pre-trained network to a new task. Sun \etal~\cite{munkhdalai2018rapid} and Ye \etal~\cite{feat_cvpr20} attempt to learn some mapping functions utilized to transform the embedding module for adaptation to a new task. Chen \etal~\cite{chen2020diversity} present a diversity transfer network that learns to generate diverse samples for novel categories, and then generate a linear classifier based on these diverse samples. Due to the generative flexibility of the meta learner, these methods achieve state-of-the-art performance in FSL~\cite{li2019lgm,chen2020diversity}. However, most of them only utilize a simple fixed metric based nearest neighbour classifier or a linear classifier for classification and thus their performance can be further improved. In this study, the generated deep metric is structured by a three-layer attentive network that is more powerful and discriminative in FSL. In addition, for metric generation, the proposed meta-learner establishes a multi-modal weight distribution condition on cross-class sample pairs, which can better exploit the task context than the uni-modal weight distribution conditioned on samples utilized in existing method~\cite{li2019lgm}.

\begin{figure*}
\setlength{\abovecaptionskip}{0pt}
\begin{center}
\vspace{-0.16cm}
\includegraphics[height=4.5in,width=7.2in,angle=0]{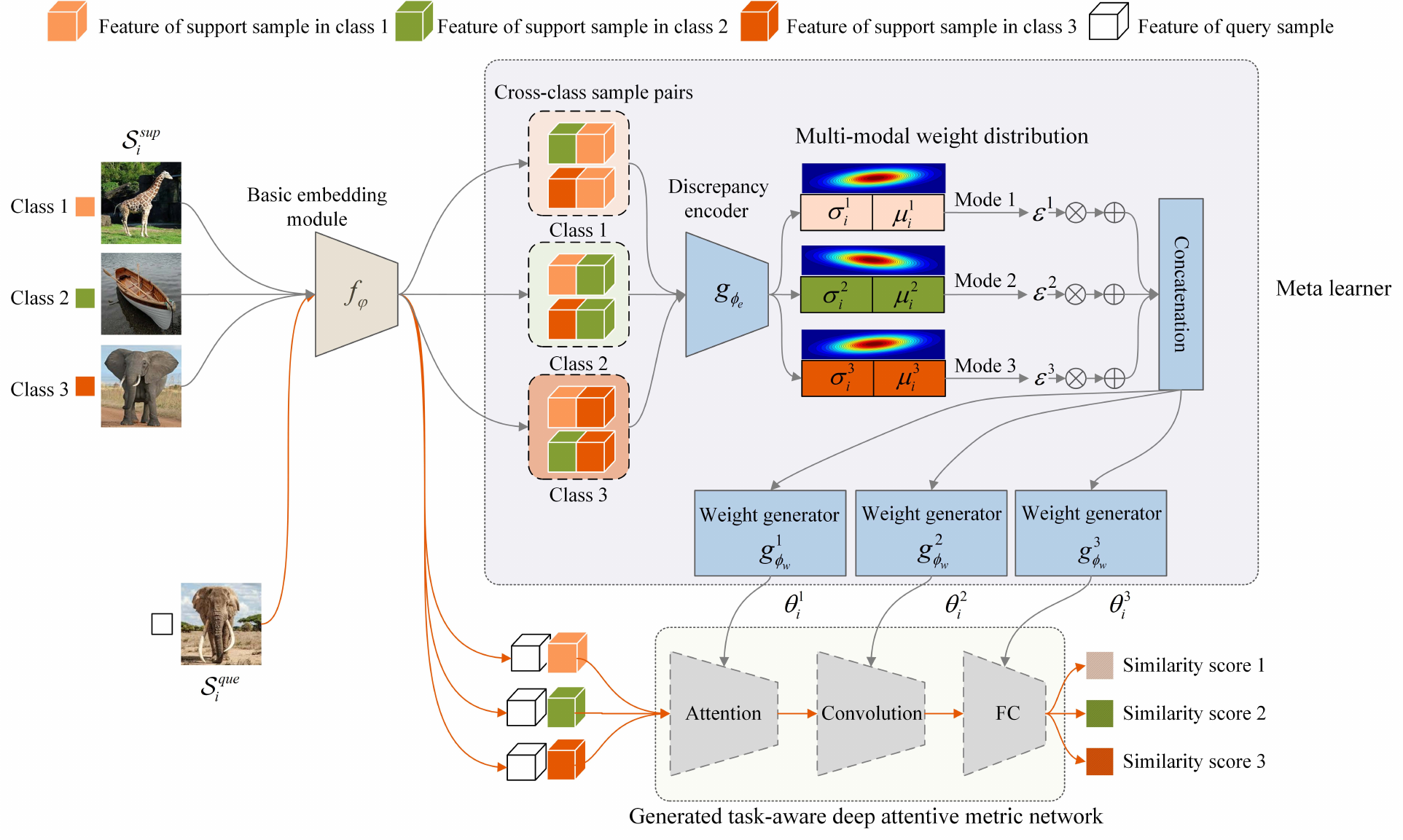}
\end{center}
\vspace{-0.4cm}
\caption{Flow chart of the proposed deep metric meta-generation method. Without loss of generality, we provide a 3-way 1-shot task with one query image. The base learner, \ie, the deep attentive metric network is generated by a meta learner that establishes a multi-modal weight distribution conditioned on the cross-class sample pairs, and produces the similarity score between the query image and each support image. $\epsilon^n$ is a random variable sampled from the standard Gaussian distribution with $n=1,2,3$. FC: fully connected layer.}
\vspace{-0.2cm}
\label{fig:flow}
\end{figure*}
 
\section{Deep Attentive Metric Meta-generation}
In this study, we formulate an $N$-way $K$-shot classification problem as that in~\cite{vinyals2016matching,li2019lgm}. Specifically, such a problem often consists of three meta datasets, including a meta training dataset $\mathcal{D}^{train}$ utilized for model training, a meta validation dataset $\mathcal{D}^{val}$ utilized for model selection and a meta test dataset $\mathcal{D}^{test}$ utilized for performance evaluation. It is noticeable that target classes in these three datasets are disjoint. With each meta dataset, we can sample a specific $N$-way $K$-shot classification task $\mathcal{T}_i$, which often contains a support set $\mathcal{S}_i^{sup}$ and a query set $\mathcal{S}_i^{que}$. The support set $\mathcal{S}_i^{sup}$ comprises $N$ classes with $K$ labelled samples per class, while the query set $\mathcal{S}_i^{que}$ contains some unlabelled samples from the same $N$ classes in $\mathcal{S}_i^{sup}$ for test.

\subsection{Methodology}
To solve the FSL task, we propose a deep attentive metric meta-generation method, which consists of three main modules, including a basic embedding module (\eg $f_\varphi$), a meta leaner and a generated task-aware deep attentive metric network, as shown in Figure~\ref{fig:flow}. The basic embedding module takes the images from both support and query sets as input and map them into a deep feature space. With the deep features of support samples, the meta learner module aims at establishing a multi-modal weight distribution conditioned on cross-class sample pairs and then sampling the distribution to generate a task-aware deep attentive metric network that is ultimately utilized to classify the query image according to its similarity scores to all support images. In the following, we will introduce the meta learner and the generated deep attentive metric in details.

\subsection{Multi-modal Weight Distribution based Meta Leaner}
It has shown~\cite{rusu2018meta,li2019lgm} that the key of generating a task-aware base learner lies on establishing a proper weight distribution conditioned on the task description (\eg, a few labelled samples) and sampling the established distribution. In this study, our aim is to generate a deep metric network. Due to the high dimensionality of the network weights, it is infeasible to directly represent such a complicated weight distribution not mentioned sampling it~\cite{kingma2019introduction}. Recent progress has shown that variational autoencoder~\cite{kingma2019introduction,pu2016variational} can utilize an encoder network to reparametrize a complicated data distribution into a latent Gaussian distribution in a deep feature space and then transform the sample of the latent Gaussian distribution using a decoder network to produce the sample of the complicated data distribution. Inspired by this, we propose a tailored variational autoencoder as the meta learner, which comprises a discrepancy encoder and several weight generators (\ie decoders), as shown in Figure~\ref{fig:flow}. Different from the standard variational autoencoder which learns to reconstruct the data fed into the encoder, the proposed meta-learner aims at transforming the input of the discrepancy encoder, \ie, task description, into a deep metric network.

\vspace{-0.3cm}
\paragraph{Discrepancy encoder} To reparametrize the conditioned weight distribution as a latent Gaussian distribution, some previous works~\cite{rusu2018meta,li2019lgm} employ an encoder network to infer the mean and standard deviation for the latent Gaussian distribution from each support sample independently. Then, the average mean and standard deviation that are computed on the results inferred from all support samples, are utilized to determine the latent Gaussian distribution. However, due to the limited amount (\eg, $N*K$) of support images, the obtained latent Gaussian distribution is inaccurate, especially when the weight distribution is complicated. Moreover, considering each sample independently makes it difficult to exploit the discriminative information between any two classes that is important for metric generation~\cite{chen2020diversity,li2020boosting}. To address these two problems with one stone, we design a discrepancy encoder to establish a multi-modal weight distribution conditioned on the cross-class sample pairs, as shown in Figure~\ref{fig:flow}. Concretely, for a given task $\mathcal{T}_i$ with a support set $\mathcal{S}^{sup}_i=\{(\mathbf{x}^{n,k}_i, y^{n,k}_i)|n=1\cdots N, k=1\cdots K\}$, where $\mathbf{x}^{n,k}_i$ denotes the feature of the $k$-th support sample from the $n$-th class obtained in the basic embedding module $f_\varphi$ in Figure~\ref{fig:flow} and $y^{n,k}_i$ denotes its label, we first concatenate each sample from the $n$-th class with any sample from the other classes along the feature dimension to produce cross-class sample pairs as
\begin{equation}\label{eq:eq1}
\begin{aligned}
\mathbf{z}^{n,j}_i = \mathtt{Concat}\left(\mathbf{x}^{n,k}_i, \mathbf{x}^{d,l}_i\right),{\kern 2pt} {\rm s.t.} {\kern 2pt} n \neq d;  k,l \in[1,K] ,
\end{aligned}
\end{equation}
where $z^{n,j}_i$ denotes the $j$-th cross-class sample pair for the $n$-th class. This is inspired by the observation that the concatenation of cross-class sample is beneficial to reveal the margin between two specific classes~\cite{chen2020diversity,li2020boosting}. Then, each cross-class sample $z^{n,j}_i$ is fed into the discrepancy encoder to infer the parameters of the latent distribution. Different from~\cite{rusu2018meta,li2019lgm} that averages out all inferred distribution parameters and obtain a uni-mode latent Gaussian distribution, we cluster all cross-class sample pairs $z^{n,j}_i$ for $j=1,...,K^2(N-1)$ as samples for the $n$-th class, and then compute the sample mean $\mu^{n}_i$ and the sample standard deviation $\sigma^{n}_i$ of the encoding results of these samples as
\begin{equation}\label{eq:eq2}
\begin{aligned}
\mathbf{\mu}^{n}_i &= \frac{1}{K^2(N-1)}\sum\limits^{K^2(N-1)}_{j=1} g_{\phi_e}\left(\mathbf{z}^{n,j}_i\right), \\
\mathbf{\sigma}^{n}_i &= \sqrt{\frac{1}{K^2(N-1)-1}\sum\limits^{K^2(N-1)}_{j=1} \left[g_{\phi_e}\left(\mathbf{z}^{n,j}_i\right) - \mathbf{\mu}^{n}_i\right]^2}.
\end{aligned}
\end{equation}
With $\mu^{n}_i$ and $\sigma^{n}_i$ computed for $N$ classes, we can obtain $N$ latent Gaussian distributions, \eg, $\mathcal{N}(\mathbf{\mu}^{n}_i, \mathbf{\sigma}^{n}_i)$. Due to being conditioned on the cross-class sample pairs, these latent $\mathcal{N}(\mathbf{\mu}^{n}_i, \mathbf{\sigma}^{n}_i)$ are able to exploit the inter-class discriminative information for metric generation. In addition, for each $\mathcal{N}(\mathbf{\mu}^{n}_i, \mathbf{\sigma}^{n}_i)$, there are $K^2(N-1)\gg K$ cross-class sample pairs utilized for distribution parameter inference, which enables modelling the distribution accurately. In this study, due to reparametrizing the conditioned weight distribution using $N$ different latent Gaussian distributions, we term it as a {\textit{multi-modal weight distribution}}. With these latent $\mathcal{N}(\mathbf{\mu}^{n}_i, \mathbf{\sigma}^{n}_i)$s, we randomly sample each of them and concatenate these produced samples as the sample $\mathbf{d}_i$ for the multi-modal weight distribution as  
\begin{equation}\label{eq:eq3}
\begin{aligned}
\mathbf{d}^n_i&= \mathbf{\epsilon} \odot \mathbf{\sigma}^n_i + \mathbf{\mu}^{n}_i,{\kern 4pt} {\rm s.t.}{\kern 4pt} \mathbf{\epsilon}\sim \mathcal{N}\left(\mathbf{0},\mathbf{I}\right),\\
\mathbf{d}_i &= \mathtt{Concat}\left(\mathbf{d}^1_i,\mathbf{d}^2_i,...,\mathbf{d}^N_i\right),
\end{aligned}
\end{equation}
where $\mathbf{d}^n_i$ denotes a random sample produced by $\mathcal{N}(\mathbf{\mu}^{n}_i, \mathbf{\sigma}^{n}_i)$ and $\mathcal{N}\left(\mathbf{0},\mathbf{I}\right)$ is the standard normal distribution.

\vspace{-0.3cm}
\paragraph{Layer-aware Weight Generator} Given the sampled result $\mathbf{d}_i$, we employ several layer-aware weight generators to separately generate the weights for each layer in the base learner. For the $m$-th layer in the base leaner, the corresponding weight can be generated as
\begin{equation}\label{eq:eq4}
\begin{aligned}
\theta^{m}_{i}=g^m_{\phi_w}\left(\mathbf{d}_i\right),
\end{aligned}
\end{equation}
where $g^m_{\phi_w}$ and $\theta^{m}_{i}$ denote the weight generator and the generated weight. To avoid producing weights with too large values and stabilize the training process, we follow~\cite{salimans2016weight,li2019lgm} and employ a weight normalization without learnable parameters. In particular, for the generated weight in a convolution layer, each kernel in the weight is normalized by its $\ell_2$ norm (\ie, $\|\cdot\|_2$) as
\begin{equation}\label{eq:eq5}
\begin{aligned}
\theta^{m,j}_{i}={\theta^{m,j}_{i}}/{\|\theta^{m,j}_{j}\|_2},
\end{aligned}
\end{equation}
where $\theta^{m,j}_{i}$ denotes the $j$-th kernel in the weight $\theta^{m,j}_{i}$. For the generated weight in a fully connected layer, the normalization is performed along the hyperplane. For the attention layer, we utilize the Sigmoid activation function~\cite{kwan1992simple} to map the generated attention into the range [0,1].

\subsection{Deep Attentive Metric}
In this study, we employ a three-layer attentive deep network to produce a discriminative metric for each task $\mathcal{T}_i$. As shown in Figure~\ref{fig:flow}, the proposed metric network consists of an attention layer, a convolution layer and a fully connected layer. For the input support and query image pair ${\mathbf{x}^{n,k}_i, \mathbf{z}^j_i}$\footnote{$\mathbf{z}^j_i$ denotes the feature of the query image obtained through the basic embedding module.}, the attention layer imposes an class-aware element-wise attention onto the concatenation of ${\mathbf{x}^{n,k}_i, \mathbf{z}^j_i}$ as
\begin{equation}\label{eq:eq6}
\begin{aligned}
\mathbf{f}^{n,k,j}_i = \mathtt{Concat}\left(\mathbf{x}^{n,k}_i, \mathbf{z}^j_i\right)\odot \left(\mathbf{a}^n_i + \mathbf{1}\right)
\end{aligned}
\end{equation}
where $\mathbf{a}^n_i$ is the attention generated by the weight generator mentioned above for the $n$-th class in task $\mathcal{T}_i$. $\mathbf{f}^{n,k,j}_i$ denotes the attentive feature. It is noticeable that such a class-aware attention is generated based on the sample $\mathbf{d}^n_i$ of the corresponding $n$-th class rather than the concatenation results $\mathbf{d}_i$ in Eq.\eqref{eq:eq5}. Then, the attentive feature $\mathbf{f}^{n,k,j}_i$ is mapped into a similarity score by the followed convolution layer and the full connected layer as
\begin{equation}\label{eq:eq7}
\begin{aligned}
s^{n,k,j}_i = \mathbf{W}^{fc}_{i}*\mathtt{ReLu}(\mathbf{f}^{n,k,j}_i \circledast \mathbf{w}_i) + b_i,
\end{aligned}
\end{equation}
where $\mathbf{w}_i$ is the generated convolution kernel. $\mathbf{W}^{fc}_{i}$ and $b_i$ are the generated weight and bias for the fully connected layer. $\mathtt{ReLu}(\cdot)$ denotes the rectified linear unit based activation function~\cite{dahl2013improving}. $s^{n,k,j}_i$ is the similarity score between the query image $\mathbf{z}^j_i$ and the support image $\mathbf{x}^{n,k}_i$. Given the similarity scores of the query image $\mathbf{z}^j_i$ to all support images, we can estimate its label as
\begin{equation}\label{eq:eq8}
\begin{aligned}
&a(z^j_i, x^{n,k}_i) = \frac{e^{s^{n,k,j}_i}}{\sum\nolimits^N_{n=1}\sum\nolimits^K_{k=1}e^{s^{n,k,j}_i}},\\
&\hat{y}^j_i = \sum\limits^N_{n=1}\sum\limits^K_{k=1} a(z^j_i, x^{n,k}_i)y^{n,k}_i.
\end{aligned}
\end{equation}

\subsection{Algorithm}
We employ the cross entropy loss $H(\cdot)$ between the predicted label and its ground truth as an objective function to train the meta learner in an end-to-end manner as 
\begin{equation}\label{eq:eq9}
\begin{aligned}
\mathcal{L}_{\mathcal{T}_i} = H(\hat{y}^j_i , y^j_i).
\end{aligned}
\end{equation}
The details can be found in Algorithm~\ref{alg:dces}.

\begin{algorithm}
\caption{Training algorithm of the proposed method for $N$-way $K$-shot tasks}
\label{alg:dces}
\KwIn{Meta training dataset $\mathcal{D}^{train}$, meta learner with parameter $\phi$, basic embedding module with parameter $\varphi$.}
{\bf{Initialization}}: Randomly initialize $\phi$ and $\varphi$.\\
\While {not converged}
{
$1.$ Sample a $N$-way $K$-shot task batch $\mathcal{T}^{batch}$ from $\mathcal{D}^{train}$\\
\For {each task $\mathcal{T}_i$ in $\mathcal{T}^{batch}$}
{
$2.$ Collect support set $\mathcal{S}^{sup}_i$ and query set $\mathcal{S}^{que}_i$;\\
$3.$ Generate weight for the deep attentive metric network, \ie, $\mathbf{a}^n_i$, $\mathbf{w}_i$, $\mathbf{W}^{fc}_i$ and $b_i$ using the mate learner based on $\mathcal{S}^{sup}_i$;\\ 	
$4.$ Compute the loss $\mathcal{L}_{\mathcal{T}_i}$ using the generated deep attentive metric on task $\mathcal{T}_i$; 
}
$5.$ Compute the batch loss $\mathcal{L}_{\mathcal{T}^{batch}}=\sum\nolimits_{\mathcal{T}_i}\mathcal{L}_{\mathcal{T}_i}$;\\
$6.$ Update $\phi$, $\varphi$ using $\nabla_\phi\mathcal{L}_{\mathcal{T}^{batch}}$ and $\nabla_\varphi\mathcal{L}_{\mathcal{T}^{batch}}$;\\
}
\KwOut{$\phi$ and $\varphi$.}
\end{algorithm}

\section{Experimental Analysis}
In this section, we conduct experiments to evaluate the efficacy of the proposed method in coping with FSL problem. First, we introduce the model architecture and experiments details of the proposed method in Section~\ref{subsec:impl}. Next, we present the experimental results on standard few-shot classification benchmarks including ImageNet derivatives in Section~\ref{subsec:imgenet} and the CIFAR derivatives in Section~\ref{subsec:cifar}, followed by an ablation study in Section~\ref{subsec:ablation}. All of the following experiments are implemented in TensorFlow~\cite{abadi2016tensorflow}. The code will be released on publication.

\subsection{Implementation Details}\label{subsec:impl}
%\vspace{-0.3cm}
\paragraph{Model Architecture} As shown in Figure~\ref{fig:flow}, the proposed method employs a basic embedding module together with the deep attentive metric generated by a meta leanrer for FSL prediction. For the basic embedding module, similar as most previous works~\cite{sung2018learning,ravi2017optimization,finn2017model,li2019lgm}, we stack four conventional blocks to extract basic features for input images. Specifically, we utilize four 3$\times$3 convolutional layers with 64 filters, followed by batch normalization (BN)~\cite{santurkar2018does}, a ReLU~\cite{dahl2013improving} non-linear activation function, and 2$\times$2 max-polling. In the meta leaner, we stack three 3$\times$3 convolutional blocks with 64 filters followed by BN, ReLU non-linear activation functions, 2$\times$2 max-polling (average pooling at the end) as the discrepancy encoder. In addition, we separately employ three single layer perceptrons as the weight generators to separately produce the attention layer, a 3$\times$3 convolution layer with 64 filters and a single fully connected layer to construct the deep attentive metric network. For simplicity, the proposed deep attentive metric meta-generation method is termed 'DAM'.

\vspace{-0.3cm}
\paragraph{Training \& test settings} In training phase, we resize all images to the same size and augment each image by random rotation. The details will be given in the following sections. For the optimization, we utilize the Adam optimizer~\cite{kingma2014adam} with beta1 0.9. Following the setting in~\cite{li2019lgm}, we train the model for 120 epochs. In each epoch, we randomly sample 1000 $N$-way $K$-shot tasks, each of which comprises $K$ training samples and 15 query samples per class. The learning rate is initialized to 1e-3, and then decayed by 0.9 every 2 epochs. In test phase, following the standard protocol utilized in~\cite{sung2018learning,lee2019meta}, we report the average classification accuracy with $\pm$95\% confidence intervals in 1000 random tasks. In every task, we use the same settings as the training phase. The test model is chosen based on the accuracy on the meta validation set.

\subsection{Experiments on ImageNet Derivatives}\label{subsec:imgenet}
%\vspace{-0.3cm}
\paragraph{\emph{mini}-ImageNet} The \emph{mini}-ImageNet dataset~\cite{vinyals2016matching} is a widely used benchmark dataset for FSL. There are 100 categories with 600 samples per category chosen from the ILSVRC-2012~\cite{russakovsky2015imagenet}, and each image is of size 84$\times$84. Following the split in~\cite{ravi2016optimization}, these categories are randomly split into 64, 16 and 20 classes for meta-training, meta-validation, and meta-testing respectively. In the experiments, we randomly rotate each image by -45, -22.5, 22.5, or 45 degrees for data augmentation as that in~\cite{li2019lgm}. Unless otherwise specified, we adopt the same data augmentation strategy for all the following datasets.

\vspace{-0.3cm}
\paragraph{\emph{tiered}-ImageNet} The \emph{tiered}-ImageNet dataset~\cite{ren18fewshotssl} is a subset of ILSVRC-2012~\cite{russakovsky2015imagenet}. There are 608 categories with 779,165 images, each category comprises a different number of images, and each image is resized to 84$\times$84. Following~\cite{vinyals2016matching}, all these classes are split into 351, 97 and 160 disjoint classes for training, validation and test respectively.

\vspace{-0.3cm}
\paragraph{Results} The numerical experiments results of different methods in various FSL tasks on the \emph{mini}-ImageNet dataset and the \emph{tiered}-ImageNet dataset are summarized in Table~\ref{table:imgnet}. As can be seen, the proposed method achieves the state-of-art performance on both datasets and surpasses all comparison methods with clear margins, especially when the task is challenging. For example, on the \emph{mini}-ImageNet dataset, the improvement of the proposed method over the other methods is even up to 20.55\% on the 20-way 1-shot task. The reason is intuitive. Compared with those metric based methods, \eg, Matching Networks~\cite{vinyals2016matching}, GNN~\cite{satorras2018few} \etc. that utilize a general comparison metric (\eg, cosine distance) for all tasks, the proposed method learns to generate a task-aware deep metric and thus generalizes better to the unseen samples in this task. Although those gradient based methods, \eg, MetaOpt-SVM~\cite{lee2019meta}, SIB~\cite{Hu2020Empirical} \etc, often performs better than those metric based ones, due to the computation of the complicated second-order gradient, most of them only learns the optimization algorithm under a linear classification rule~\cite{lee2019meta,zhang2020deepemd}. In contrast, the proposed method can learn to generate a deep discriminative metric in an end-to-end training manner. For these generation based method, while also learning to generate a base learner, they mainly focus on generating an task-aware embedding module and then predicting with a fixed metric (\eg LGM-net~\cite{li2019lgm} AdaResNet~\cite{munkhdalai2018rapid} \etc), or directly generating a linear classifier (\eg, Deep DTN~\cite{chen2020diversity}). Their performance is limited by the fixed metric or the simple classier. In contrast, the proposed method generates a deep discriminative metric for prediction. Although TADAM~\cite{oreshkin2018tadam} also learns a task-aware metric, it only learns a scaling factor for the cosine distance metric. In addition to the accuracy, the backbone complexity of the proposed method is also moderate among all methods. Thus, the proposed method s arguably simpler and achieves strong performance.

\begin{table*}\footnotesize%\scriptsize%
	\caption{Evaluation results of different methods on \emph{mini}-ImageNet and \emph{tiered}-ImageNet datasets. Average classification accuracies (\%) with 95\% confidence intervals on the meta-test split are provided. In the backbone column, 'a-b-c-d' denotes a 4-layer convolutional network with a, b, c, and d filters in each layer. The comparison methods are divided into three categories, including the metric based methods, gradient based methods and the generation based methods, from top to bottom. The best results are in bold.}
\vspace{-0.3cm}
\label{table:imgnet}
\renewcommand{\arraystretch}{1.15}
\begin{center}
\scalebox{0.95}{
	\begin{tabular}{lll|ccc|ccc}
		\toprule[1.2pt]
		 & & & \multicolumn{3}{c|}{\emph{mini}-ImageNet} & \multicolumn{2}{c}{\emph{tiered}-ImageNet}
		 \\
		 \hline
		{\textbf{Model}} & 
		{\textbf{mark}} & 
		{\textbf{backbone}} &
		{\textbf{5-way 1-shot}} & 
		{\textbf{5-way 5-shot}} & 
		{\textbf{20-way 1-shot}}&
		{\textbf{5-way 1-shot}} & 
		{\textbf{5-way 5-shot}} & 
		\\
		\toprule[0.6pt]
		Matching Networks~\cite{vinyals2016matching} & NeurIPS'16 & 64-64-64-64 &	
		43.56$\pm$0.84 & 
		55.31$\pm$0.73 &
		17.31$\pm$0.22 &
		- & -\\
		Prototypical Networks~\cite{snell2017prototypical} & NeurIPS'17 & 64-64-64-64 & 
		49.42$\pm$0.78 &
	    68.20$\pm$0.66 &
	    -&
	    53.31$\pm$0.89 &
		72.69$\pm$0.74\\
		Relation Networks~\cite{sung2018learning} & CVPR'18 & 64-96-128-256 & 
		50.44$\pm$0.82 & 
		65.32$\pm$0.70 &
		-&
		54.48$\pm$0.93 &
		71.32$\pm$0.78\\
		Transductive Prop Nets~\cite{liu2018learning} & ICLR'19 & 64-64-64-64 & 
		55.51$\pm$0.86 & 
		69.86$\pm$0.65 &
		-&
		59.91$\pm$0.94 &
		73.30$\pm$0.75\\
		GNN~\cite{satorras2018few} & ICLR'18 & 
		64-64-64-64 & 
		50.3 & 
		66.4 &
		- & - & -\\
		RFS~\cite{tian2020rethink} & ECCV'20 & 
		ResNet-12 &
		64.82$\pm$0.60 &
		82.14$\pm$0.43 &
		-&
		71.52$\pm$0.69 &
	    86.03$\pm$0.49		
		\\
		\hline
		MAML~\cite{finn2017model} & ICML'17 & 
		32-32-32-32 & 
		48.70$\pm$1.84 & 
		63.11$\pm$0.92 &
		16.49$\pm$0.58 &
		51.67$\pm$1.81 &
		70.30$\pm$1.75  
		\\
		Meta-Learning LSTM~\cite{ravi2017optimization}& ICLR'17 & 64-64-64-64 & 
		43.44$\pm$0.77 & 
		60.60$\pm$0.71 &
		16.70$\pm$0.23 & 
		- & -\\
		R2D2~\cite{bertinetto2018meta} & ICLR'19 & 
		96-192-384-512 & 
		51.2$\pm$0.6 & 
		68.8$\pm$0.1 & 
		- & - & -\\
		LEO~\cite{rusu2018meta} & ICLR'19 & 
		WRN-28-10 & 
		61.76$\pm$0.08 & 
		77.59$\pm$0.12 &
		-&
		66.33$\pm$0.05 &
		81.44$\pm$0.09		
		\\
		MetaOptNet-RR~\cite{lee2019meta} & CVPR'19 & 
		ResNet-12 & 
		61.41$\pm$0.61 & 
		77.88$\pm$0.46 &
		-&
		65.36$\pm$0.71 &
		81.34$\pm$0.52		
		\\
		MetaOptNet-SVM~\cite{lee2019meta} & CVPR'19 & 
		ResNet-12 & 
		62.64$\pm$0.61 & 
		78.63$\pm$0.46 &
		-&
	    65.99$\pm$0.72 &
	    81.56$\pm$0.53		
		\\
		SIB~\cite{Hu2020Empirical} & ICLR'20 & 
		WRN-28-10 & 
		70.0$\pm$0.6 & 
		79.2$\pm$0.4 &
		- & - & -\\
		DeepEMD~\cite{zhang2020deepemd} & CVPR'20 & 
		ResNet-12 & 
		65.91$\pm$0.82 &
		82.41$\pm$0.56 &
		-&
		71.16$\pm$0.87 &
		86.03$\pm$0.58 \\		
		\hline
%		SNAIL~\cite{mishra2017simple} & ICLR'18 & 
%		ResNet-12 & 
%		55.71$\pm$0.99 & 
%		68.88$\pm$0.92 &
%		- & - & -\\
		AdaResNet~\cite{munkhdalai2018rapid} & ICML'18 & ResNet-12 & 
		56.88$\pm$0.62 & 
		71.94$\pm$0.57 &
		- & - & -\\
%		Activation to Parametery~\cite{Qiao_2018_CVPR} & CVPR'18 & WRN-28-10 & 
%		59.60$\pm$0.41 &
%	    73.74$\pm$0.19 &
%	    - & - & -\\
	    TADAM~\cite{oreshkin2018tadam} & NeurIPS'18 & 
		ResNet-12 & 
		58.50$\pm$0.30 & 
		76.70$\pm$0.30 &
		- & - & -\\
		LGM-net~\cite{li2019lgm} & ICML'19 & 
		64-64-64-64 & 
		69.13$\pm$0.35 & 
		71.18$\pm$0.68 &
		26.14$\pm$0.34 &
		- & - & -\\
		Dynamic Few-shot~\cite{Gidaris_2018_CVPR} & CVPR'18 & 64-64-128-128 & 
		56.20$\pm$0.86 & 
		73.00$\pm$0.64 &
		- & - & -\\
		Deep DTN~\cite{chen2020diversity} & AAAI'20 & ResNet-12 & 63.45$\pm$0.86 & 77.91$\pm$0.62
		& - & - & -\\
		FEAT~\cite{feat_cvpr20} & CVPR'20 & 
		WRN-28-10 & 
		65.10$\pm$0.20 & 
		81.11$\pm$0.14 &
		-&
		70.41$\pm$0.23 &
		84.38$\pm$0.16		
		\\
%		LaplacianShot~\cite{laplacianshot20} & 
%		ResNet-18 & 
%		72.11$\pm$0.19 & 
%		82.31$\pm$0.14 &
%		-\\
		\hline
		DAM (ours) & - & 
		64-64-64-64 & 
		{\textbf{79.15$\pm$0.35}} & 
		{\textbf{87.75$\pm$0.36}} &
		{\textbf{46.69$\pm$0.36}} &
		{\textbf{78.63$\pm$0.38}} &
		{\textbf{87.79$\pm$0.36}}
		\\
		\bottomrule[1.2pt]
		\end{tabular}
		}
	\end{center}
	\vspace{-0.5cm}
\end{table*}

\subsection{Experiments on CIFAR Derivatives}\label{subsec:cifar}
%\vspace{-0.2cm}
\paragraph{CIFAR-FS} CIFAR-FS dataset ~\cite{bertinetto2018meta} is derived from CIFAR100~\cite{krizhevsky2009learning}, which splits all classes in CIFAR100 into 64, 16 and 20 classes for training, validation, and testing respectively. There are 600 samples per class and the resolution of every image is 32$\times$32.

\vspace{-0.3cm}
\paragraph{FC100} FC100 dataset ~\cite{oreshkin2018tadam} is also originated from CIFAR100. There are 60, 20 and 20 classes for training, validation, and testing respectively. The resolution and the number of samples per class are the same as CIFAR-FS. It is worth noting that, compared with CIFAR-FS, FC100 has a lower information overlap in the partition of training set, validation set and test set, which makes it more challenging.

\vspace{-0.3cm}
\paragraph{Results} The numerical results of different methods in various tasks on both CIFAR-FS and FC100 datasets are provided in Table~\ref{table:cifar}. As can be seen, the proposed method also outperforms all comparison methods with clear margins, especially on the more challenging FC100 where the improvement over the other methods is up to 24.12\% and 17\% in the 5-way 1-shot task and 5-way 5-shot, respectively.

\begin{table*}\footnotesize%
	\caption{Evaluation results of different methods on CIFAR-FS and FC100 datasets. Average classification accuracies (\%) with 95\% confidence intervals on the meta-test split are provided. In the backbone column, 'a-b-c-d' denotes a 4-layer convolutional network with a, b, c, and d filters in each layer. The comparison methods are divided into three categories, including the metric based methods, gradient based methods and the generation based methods, from top to bottom. The best results are in bold.}
\vspace{-0.3cm}
	\label{table:cifar}
	\renewcommand{\arraystretch}{1.15}
	\begin{center}
	\scalebox{1.0}{
		\begin{tabular}{lll|cc|cc}
			\toprule[1.2pt]
			&&& \multicolumn{2}{c|}{CIFAR-FS} & \multicolumn{2}{c}{FC100}
			\\
			\hline			
			{\textbf{Model}} & 
			{\textbf{mark}} & 
			{\textbf{backbone}} & 
			{\textbf{5-way 1-shot}} & 
			{\textbf{5-way 5-shot}} &
			{\textbf{5-way 1-shot}} & 
			{\textbf{5-way 5-shot}}\\
			\toprule[0.6pt]
			Prototypical Networks~\cite{snell2017prototypical} & NeurIPS'17 & 64-64-64-64 & 
			55.5$\pm$0.7 &
			72.0$\pm$0.6 &
			35.3$\pm$0.6 &
			48.6$\pm$0.6 
			\\
			Relation Networks~\cite{sung2018learning} & CVPR'18 & 64-96-128-256 & 
			55.0$\pm$1.0 &
			69.3$\pm$0.8 & - & -\\
			GNN~\cite{satorras2018few} & ICLR'18 & 
			64-64-64-64 & 
			61.9 & 75.3 & - & - \\
			RFS~\cite{tian2020rethink} & ECCV'20 & 
			ResNet-12 &
			73.9$\pm$0.8 &
			86.9$\pm$0.5 &
			44.6$\pm$0.7 &
		    60.9$\pm$0.6 \\
			\hline
			MAML~\cite{finn2017model} & ICML'17 & 
			32-32-32-32 & 
			58.9$\pm$1.9 &
			71.5$\pm$1.0 &
			38.1 & 50.4 
			\\
			R2D2~\cite{bertinetto2018meta} & ICLR'19 & 
			96-192-384-512 & 
			62.3$\pm$0.2 &
			77.4$\pm$0.2 & - & -\\
			MetaOptNet-RR~\cite{lee2019meta} & CVPR'19 & 
			ResNet-12 & 
			72.6$\pm$0.7 &
			84.3$\pm$0.5 &
			40.5$\pm$0.6 &
			55.3$\pm$0.6 \\
			MetaOptNet-SVM~\cite{lee2019meta} & CVPR'18 & 
			ResNet-12 & 
			72.0$\pm$0.7 &
			84.2$\pm$0.5 &
			41.1$\pm$0.6 &
		    55.5$\pm$0.6 \\
		    %SIB~\cite{Hu2020Empirical} & ICLR'20 & 
		    %64-64-64-64 & 
		    %68.7$\pm$0.6 &
		    %77.1$\pm$0.4 \\
		    SIB~\cite{Hu2020Empirical} & ICLR'20 & 
		    WRN-28-10 & 
		    80.0$\pm$0.6 &
		    85.3$\pm$0.4 &
		    45.2 & 55.9 \\
			\hline
			TADAM~\cite{oreshkin2018tadam} & NeurIPS'18 & 
			ResNet-12 & 
			- & - &
			40.1$\pm$0.4 & 56.1$\pm$0.4
			\\ 
			\hline
			DAM (ours) & - & 
			64-64-64-64 & 
			{\textbf{80.78$\pm$0.33}} &
			{\textbf{89.80$\pm$0.32}} &
			{\textbf{68.72$\pm$0.40}} &
			{\textbf{77.92$\pm$0.44}} \\
			\bottomrule[1.2pt]
		\end{tabular}
		}
	\end{center}
	\vspace{-0.3cm}
\end{table*}

\subsection{Ablation Study}\label{subsec:ablation}
The proposed method has two key components, including the meta-learner that establishes a multi-modal weight distribution conditioned on cross-class sample pairs and the generated task-ware deep attentive metric. To demonstrate the effectiveness of each component in coping with FSL problem, without loss of generality, we conduct ablation study on the 5-way 1-shot task on the \emph{mini}-ImageNet dataset.

\begin{table}\footnotesize%
	\caption{Ablation study of the proposed method on the \emph{mini}-ImageNet dataset. The best results are in bold.}
	\vspace{-0.3cm}
	\label{table:ablation}
	\renewcommand{\arraystretch}{1.15}
	\begin{center}
	\scalebox{1.0}{
		\begin{tabular}{lc}
			\toprule[1.2pt]
			Method & 5-way 1-shot\\
			\hline
			DAM w/o cross-class sample pairs & 78.53 $\pm$ 0.39\\
			DAM w/o data variance & 78.39 $\pm$ 0.38\\
			DAM w/o multi-modal weight distribution & 78.30 $\pm$ 0.34\\
			\hline
			DAM with matching net & 68.33 $\pm$ 0.44\\
			DAM w/o attention & 78.60 $\pm$ 0.37\\
			\hline
			DAM & {\textbf{79.15 $\pm$ 0.35}}\\
			\bottomrule[1.2pt]
		\end{tabular}
		}
	\end{center}
	\vspace{-0.3cm}
\end{table}

\begin{figure*}
\setlength{\abovecaptionskip}{0pt}
\begin{center}
\vspace{-0.16cm}
\subfigure[Attention layer]{\includegraphics[height=1.0in,width=1.35in,angle=0]{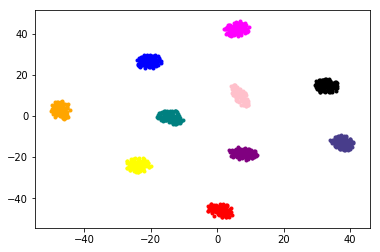}}
\hspace{-0.2cm}
\subfigure[Convolution layer]{\includegraphics[height=1.0in,width=1.35in,angle=0]{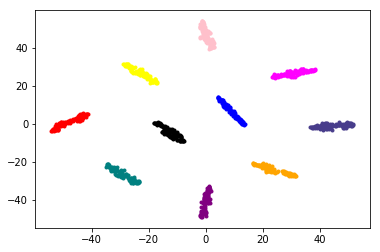}}
\hspace{-0.2cm}
\subfigure[Fully connected layer]{\includegraphics[height=1.0in,width=1.35in,angle=0]{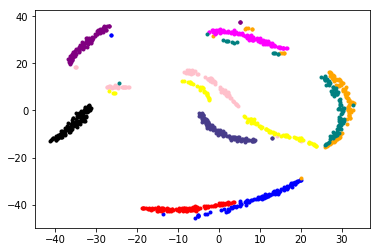}}
\hspace{-0.2cm}
\subfigure[Class-aware attention]{\includegraphics[height=1.0in,width=1.35in,angle=0]{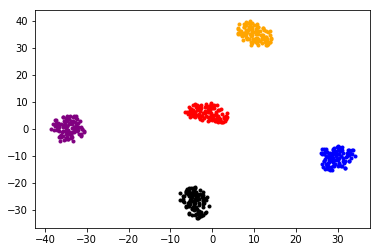}}
\hspace{-0.2cm}
\subfigure[Multi-model distribution]{\includegraphics[height=1.0in,width=1.35in,angle=0]{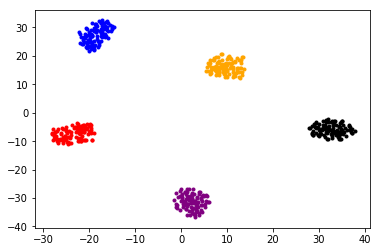}}
\end{center}
\vspace{-0.4cm}
\caption{Visualization of weight distribution and multi-model distribution. (a)-(c) gives the weight distribution of the attention layer, convolution layer and fully connected in the generated task-aware deep attentive metric network for 10 different tasks. (d) provides the weight distribution for the class-aware attention layers utilized for five different classes in a specific task. (e) shows five latent Gaussian distributions utilized for the multi-model weight distribution in the proposed meta learner.}
\vspace{-0.2cm}
\label{fig:weight}
\end{figure*}

\vspace{-0.3cm}
\paragraph{Multi-modal Parameter Distribution based Meta Learner} In the proposed meta learner, it is necessary to separately discuss the efficacy of the usage of the cross-class sample pairs, the computation of standard deviation in Eq.~\eqref{eq:eq2} as well as the design of multi-modal weight distribution. To demonstrate the efficacy of the usage of cross-class sample pairs, we implement a variant 'DAM w/o cross-class sample pairs' that takes each support image as an independent sample to infer the parameters of a latent uni-modal Gaussian distribution as that in LGM-net~\cite{Gidaris_2018_CVPR} while keep the other settings unchanged. In addition, to show the effect of the computation of the standard deviation in Eq.~\eqref{eq:eq2}, we implement a variant 'DAM w/o data variance' by replacing the computation of the standard deviation in Eq.~\eqref{eq:eq2} by directly inferring it as the mean value in Eq.~\eqref{eq:eq2} using the discrepancy encoder as that in LGM-net~\cite{Gidaris_2018_CVPR}. To demonstrate the efficacy of the design of multi-modal weight distribution, we implement a variant 'DAM w/o multi-model weight distribution' by putting all cross-class sample pairs together and computing a global mean and standard deviation that are shared by all classes. By doing this, the multi-modal weight distribution degenerates to a uni-modal distribution. Table~\ref{table:ablation} summarizes the numerical results of all these variants. As can be seen, the performance of these variants is inferior to the proposed DAM, which respectively demonstrates that the usage of cross-class sample pairs, the computation of the standard deviation in Eq.~\ref{eq:eq2} and the multi-model weight distribution do matter for the proposed method. This is because the cross-class sample pairs are the key to exploit the inter-class discrepancy using the discrepancy encoder, while the multi-model weight distribution can exhaustively exploit the class-wise inter-class discrepancy statistics and summarize them for metric generation. The computation in Eq.~\eqref{eq:eq5} can sufficiently and explicitly utilize the statistics of the cross-class sample pairs, which enables the generated metric more suitable to the task.

\vspace{-0.3cm}
\paragraph{Generated Deep Attentive Metric} To demonstrate that generating a deep metric as this study is more effective than generating an embedding module, we utilize the proposed meta learner to generate a three-layer task-aware embedding module. With the feature output by the generated embedding module, we utilize a fixed cosine distance metric for classification as that in Eq.~\eqref{eq:eq8}. For simplicity, we term this variant as 'DAM with matching net'. In addition, we also implement a variant 'DAM w/o attention module' that removes the generated attention module in the deep metric network in Figure~\ref{fig:flow}. The numerical results of these two variants are provided as the second part of Table~\ref{table:ablation}. As can be seen, the proposed method outperforms both variants with clear margins, especially the 'DAM with matching net'. This demonstrates that generating a deep metric is the key to improve the performance of FSL and the attention module can also lead to improvement at some extent.

\subsection{Visualization Results}
To better understand the proposed method, we provide some visualization results of the generated task-aware metric and the multi-modal weight distribution. Without loss of generality, we obtain these results in the 5-way 1-shot task on \emph{mini}-ImageNet dataset.
\vspace{-0.3cm}
\paragraph{Task-aware Metric Network} For ten randomly sampled tasks, we randomly generated 100 deep metrics for each task through sampling the weight distribution using the proposed meta learner. Then, we represent the weights in each layer of the metric network as a functional weight point as that in~\cite{li2019lgm} and draw them in Figure~\ref{fig:weight}(a)-(c), where the functional weight points from the same task are shown in the same color. Since we separately generate an attention layer for each class as shown in Eq.~\eqref{eq:eq6}, we concatenate the weights of all attention layer for visualization. Similarly, we also concatenate the weight and bias in the fully connected layer for visualization. As can be seen, for each layer in the metric network, different tasks are assigned differently distributed weights. %In addition, the task difference in the fully connected layer is smaller than that in the attention or the convolution layer, since the fully connect layer in different tasks will do the same thing,\ie, transform the feature into a similarity score.
To further clarify the attention layer, we fix a task and visualize the weight in every class-aware attention layer in different colors in Figure~\ref{fig:weight}(d). As can be seen, different classes are given different attention weights.

\vspace{-0.3cm}
\paragraph{Multi-model Weight Distribution} To show the weight distribution established by the meta-learner is multi-modal, for a fixed task, we visualize these five latent $\mathcal{N}(\mu^n_i,\sigma^n_i)$ defined in Eq.~\eqref{eq:eq2} in Figure~\ref{fig:weight}(e). As can be seen, these latent Gaussian distributions distribute differently, which demonstrate that each class has its specific inter-class discrepancy statistics. Thus, it is necessary to utilize a multi-modal weight distribution as this study.

\section{Conclusion}
In this study, we present a deep metric meta-generation method that learns to adaptively generate a specific metric for a new FSL task based on the task description. For the meta learner, we utilize a tailored variational autoencoder that takes advantage of cross-class sample pairs to establish a multi-modal weight distribution for metric generation. With such a meta learner, we generate a three-layer attentive metric network as the base learner. Since the metric is specially designed for a new task based on its description, the proposed method can well generalize and achieves pleasing performance on a new task. Experiments on four benchmark datasets demonstrate its efficacy. 

{\small
\bibliographystyle{ieee_fullname}
\bibliography{egbib}
}

\end{document}